\documentclass[10pt,twocolumn,letterpaper]{article}

\usepackage[pagenumbers]{cvpr} 

\usepackage[dvipsnames]{xcolor}

\definecolor{cvprblue}{rgb}{0.21,0.49,0.74}
\usepackage[pagebackref,breaklinks,colorlinks,allcolors=cvprblue]{hyperref}
\usepackage{amsmath}
\usepackage{xcolor,colortbl}
\usepackage{multirow}
\usepackage{subcaption}
\usepackage{stfloats}
\def\paperID{8941} 
\def\confName{CVPR}
\def\confYear{2025}

\newcommand{\ti}[1]{#1^i_t}

\title{HaWoR: World-Space Hand Motion Reconstruction from Egocentric Videos}

\author{Jinglei Zhang$^1$, Jiankang Deng$^2$, Chao Ma$^1$, Rolandos Alexandros Potamias$^2$ \\
 $^1$Shanghai Jiao Tong University, $^2$Imperial College London\\
{\tt\small \{{zhangjinglei168, chaoma}\}@sjtu.edu.cn}, {\tt\small \{{j.deng16, r.potamias}\}@imperial.ac.uk}
}

\begin{document}
\twocolumn[{
\renewcommand\twocolumn[1][]{#1}%
\maketitle
\begin{center}
    \centering
    \captionsetup{type=figure}
    \includegraphics[width=\textwidth]{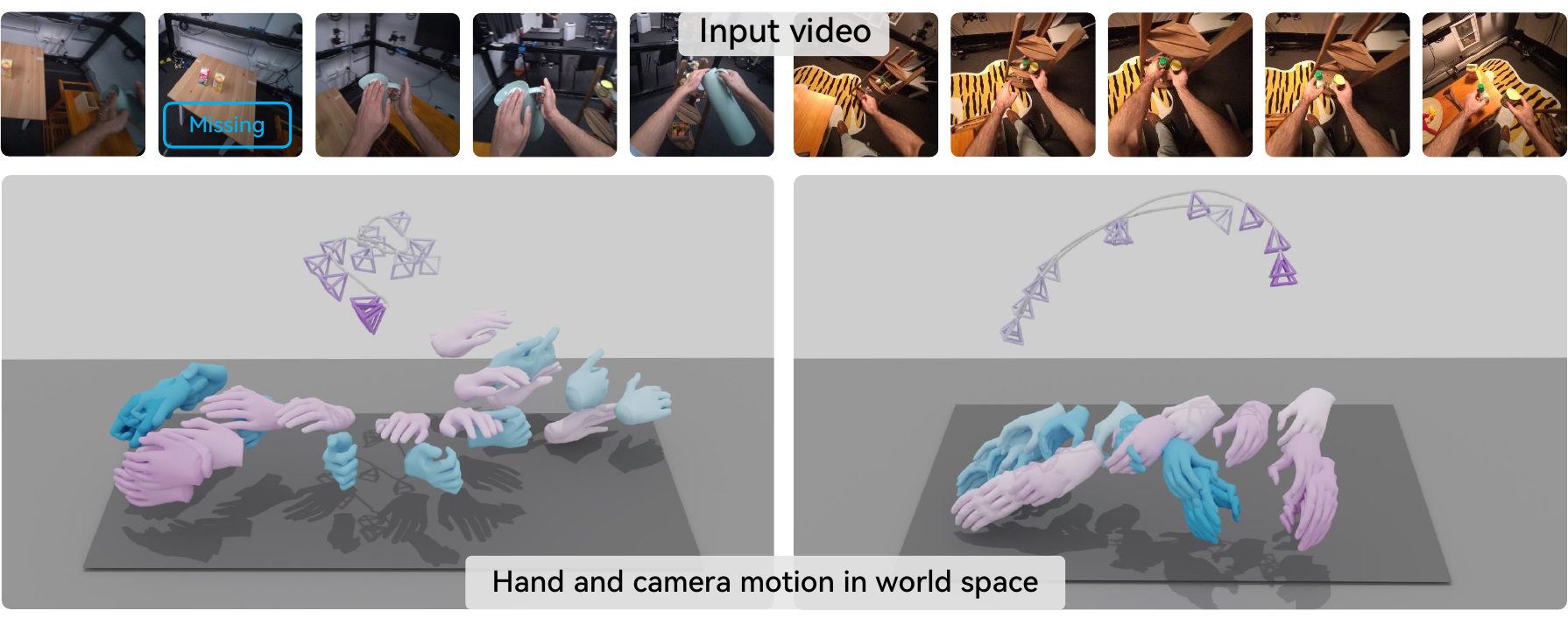}
    \captionof{figure}{We propose \textbf{HaWoR}, a world-space 3D hand motion estimation method for egocentric videos. We decouple world-space hand motion estimation by combining camera-frame motions and world-space camera trajectories. HaWoR achieves state-of-the-art performance on both camera pose estimation and hand motion reconstruction, even under challenging cases where hands are out of the view frustum.}
\end{center}}]
\maketitle
\definecolor{TableDarkGreen}{RGB}{182,215,168}
\definecolor{TableLightGreen}{RGB}{207,234,215}
\definecolor{TableYellow}{RGB}{255,242,204}
\definecolor{TableRed}{RGB}{244,204,204}

\begin{abstract}
Despite the advent in 3D hand pose estimation, current methods predominantly focus on single-image 3D hand reconstruction in the camera frame, overlooking the world-space motion of the hands.
Such limitation prohibits their direct use in egocentric video settings, where hands and camera are continuously in motion. 
In this work, we propose HaWoR, a high-fidelity method for hand motion reconstruction in world coordinates from egocentric videos. 
We propose to decouple the task by reconstructing the hand motion in the camera space and estimating the camera trajectory in the world coordinate system.  
To achieve precise camera trajectory estimation, we propose an adaptive egocentric SLAM framework that addresses the shortcomings of traditional SLAM methods, providing robust performance under challenging camera dynamics.
To ensure robust hand motion trajectories, even when the hands move out of view frustum, we devise a novel motion infiller network that effectively completes the missing frames of the sequence. 
Through extensive quantitative and qualitative evaluations, we demonstrate that HaWoR achieves state-of-the-art performance on both hand motion reconstruction and world-frame camera trajectory estimation under different egocentric benchmark datasets.
Code and models are available on our \href{https://hawor-project.github.io/}{project page}. 

\end{abstract}    
\section{Introduction}
\label{sec:intro}
Recovering fine-grained 3D hand motion estimation from monocular videos has garnered significant attention, given its critical role in various applications such as augmented/virtual reality (AR/VR) and human behavior analysis~\cite{qi2024computer,Baltatzis_2024_CVPR,bansal2024hoiref}.  
Despite the progress of 3D hand pose estimation from monocular images and videos ~\cite{lin2021end,lin2021mesh,hampali2022keypoint,hamer,potamias2024wilor}, existing approaches predominantly focus on camera-space reconstructions, often overlooking the hands' trajectories in world-space. 
Neglecting the camera motion restricts the ability of hand reconstruction methods to accurately interpret the human movements, posing a significant burden in advancing the understanding of human actions.

Estimating hand motion on world coordinates and capturing the global motion trajectory in dynamic environments is non-trivial. 
This is particularly pronounced in egocentric scenarios, where both the hands and the camera are simultaneously in motion, complicating the estimation of the scale of hand movements, resulting in trajectories that fail to reflect the true motion in world coordinates. 
Finding a direct mapping between egocentric videos and 3D world coordinates of the hands is extremely challenging due to frequent occlusions, rapid hand movements, and the dynamic interactions between the hands and the surrounding environment~\cite{banerjee2024introducing}. 
In particular, in contrast to human motion recovery, reconstructing hand motion poses challenges for two reasons. 
Firstly, the scale of hand trajectories in egocentric views is inherently more complex compared to third-person perspectives. Secondly, in egocentric scenarios, hands frequently fall outside the field of view or experience severe occlusions, making motion estimation particularly challenging. 
While human motion estimation can benefit from the use of motion priors, developing such priors of the hand motion is non-trivial due to the intricate nature of hand displacements and articulations, compounded by the limited availability of large-scale hand mocap datasets.

Early approaches in world-space human mesh recovery depend on multi-view camera setups and visual odometry systems, which often struggle to generalize beyond controlled capture environments \cite{park20153d,von2018recovering}. 
 Although Simultaneous Localization and Mapping (SLAM) methods \cite{droid_slam} have made considerable strides in capturing unstructured environments with dynamic camera movements, they often struggle when dealing with dynamic scenes that involve complex human motions.
To tackle this, several methods have approached world-frame reconstruction by leveraging heavy optimization schemes to align the human motion to estimated SLAM camera trajectories \cite{glamr,slahmr}. 
To alleviate the costly optimization process, follow-up works have attempted to utilize camera-space motion recovery methods and directly predict the camera-to-world transform \cite{shin2024wham,wang2024tram}.

Given that accurately reconstructing 3D hand motions in the world-coordinate system is significantly challenging, we propose to decompose the problem into two simpler tasks: the 3D hand motion reconstruction in the camera space and the camera trajectory estimation in the world space. 
For the first task, we train a high-fidelity transformer-based 3D hand motion reconstruction model to effectively capture hand motions in the camera space. 
However, reconstructing the 3D hand motions from egocentric videos poses significant challenges, especially when the hands are not visible within the camera frame or face severe occlusions. 
To address this, we enhance our proposed 3D hand motion reconstruction framework with a novel motion infilling module that estimates the missing and occluded hands. 
To reconstruct the camera trajectory in the world-coordinate system, we follow a hybrid method that adapts the estimated camera trajectory derived from monocular DROID-SLAM \cite{droid_slam} to the world space using a metric foundational model \cite{metric3d}.  
Nevertheless, directly using the DROID-SLAM method and the estimated world-scale from metric networks to adjust the camera trajectory leads to faulty camera trajectories that do not accurately represent the true scale of the environment. 
We effectively overcome this by proposing an adaptive version of DROID-SLAM that excludes the hand regions from the bundle adjustment state and achieves accurate and robust camera trajectories from egocentric videos. 
Similarly, we propose a normalization of the metric space to achieve accurate world scales. 

To sum up, in this paper, we present HaWoR, a robust method for 3D hand motion estimation in world coordinates from single, in-the-wild video. Specifically: 
\begin{itemize}
    \item We propose the first, to the best of our knowledge, 3D hand motion estimation method in the world-coordinate system. In contrast with previous methods that tackle 3D hand pose estimation in the camera space, we model 3D human hands in the global space, making a significant step towards real-world 3D hand motion reconstruction. 
    \item The proposed hand motion reconstruction method leverages a novel infiller network and is able to capture high-fidelity hand motions even from videos with missing frames and severe occlusions. 
    \item Finally, we propose a robust single-shot camera trajectory estimation pipeline tailored to egocentric videos, which achieves state-of-the-art performance compared to greedy optimization-based methods. 
\end{itemize}

\begin{figure*}[t]
  \centering
  \includegraphics[width=\textwidth]{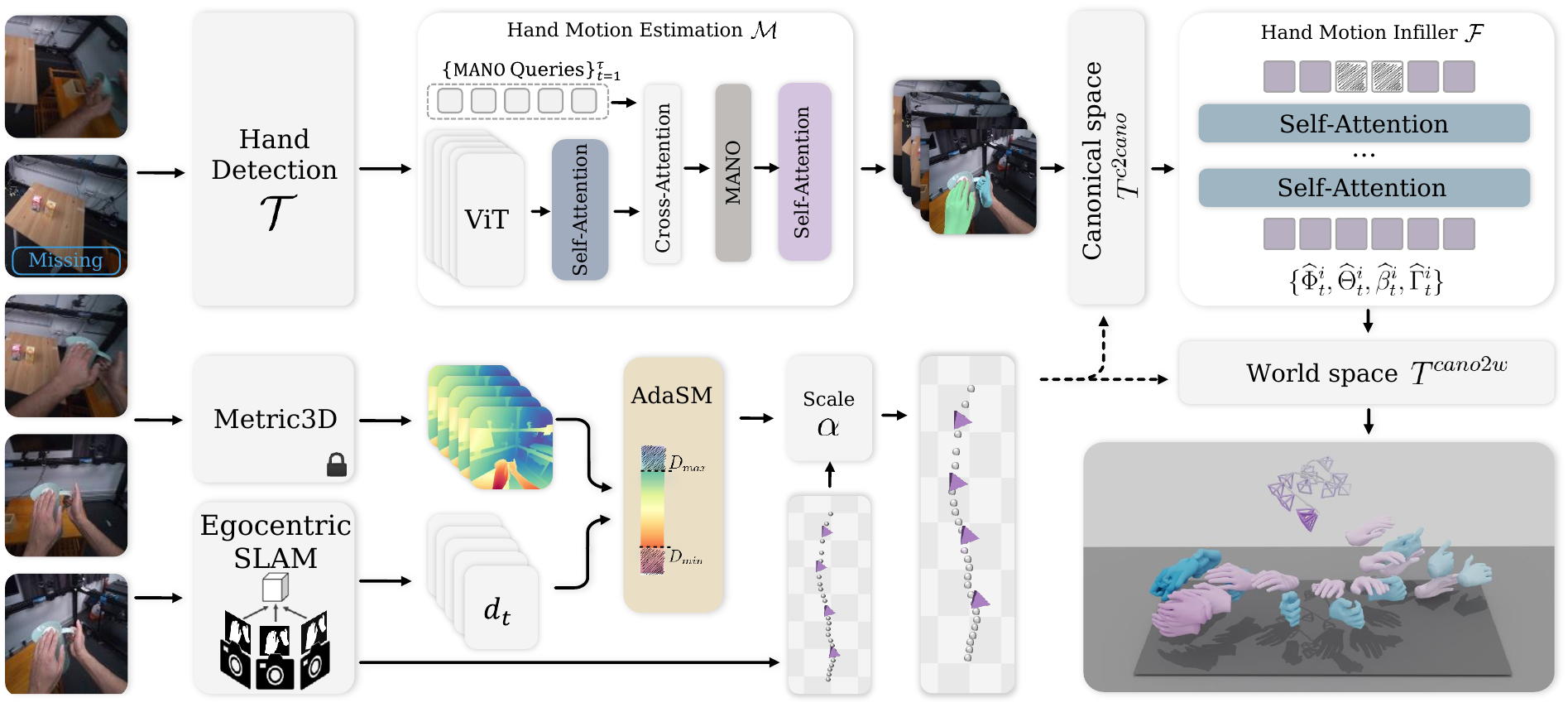}
  \caption{\textbf{Overview of our method.} Given an egocentric video $\mathbf{V}$ with a set of detected hands from an off-the-shelf detector~\cite{potamias2024wilor}, we utilize a large-scale transformer-based module with two levels of data-driven motion priors to reconstruct the 3D hand motions in the camera frame. To reconstruct hand movements beyond the view frustum, we introduce a novel hand motion infiller network designed to complete the missing frames in the hand motion sequence.
  We estimate world-space camera trajectories using an adaptive egocentric SLAM module that is accompanied by a foundation metric model \cite{metric3d} to accurately align the SLAM reconstructions to the world-coordinates. }
  \label{fig:method}
\end{figure*}

\section{Related Work}
\label{sec:related}
\noindent\textbf{3D Hand Pose Estimation} 
Hand pose estimation has been widely studied for over than a decade, where early methods utilized depth cameras to reconstruct the 3D hand~\cite{oikonomidis2011efficient,tagliasacchi2015robust,ge2016robust}. 
In the pioneering work of Boukhayma \etal~\cite{boukhayma20193d}, the authors proposed the first single-image 3D hand reconstruction method trained to estimate the hand parameters of MANO model~\cite{romero2017embodied}. 
Several methods have followed~\cite{boukhayma20193d} by regressing MANO parameters~\cite{zhang2019end,baek2019pushing} or directly predicting the 3D hand vertices~\cite{kulon2019single,lin2021end,lin2021mesh}. 
Recently, the importance of data and model scaling has been extensively highlighted, with large-scale transformer models being introduced to enhance reconstruction quality.
In particular, Pavlakos \etal~\cite{hamer} demonstrated that by utilizing a pretrained large-scale Vision Transformer (ViT) and scaling the data can effectively improve the performance. 
Potamias \etal~\cite{potamias2024wilor} introduced a refinement mechanism to progressively deforms the estimated hand pose resulting in state-of-the-art 3D hand pose estimations with accurate image-alignment. 

\noindent\textbf{3D Body and Camera Reconstruction.} Estimation of body and camera trajectory in world coordinate system was initially approached using multi-camera setup~\cite{joo2015panoptic} or additional wearable devices (e.g., IMU~\cite{von2018recovering} or electromagnetic sensors~\cite{kaufmann2023emdb}).  
GLAMR~\cite{glamr} introduced the first method for estimating global human trajectories from monocular videos with dynamic cameras, using a global trajectory regressor to infer the overall human trajectories from localized body movements. 
Several methods ~\cite{slahmr,kocabas2024pace} proposed to decouple camera and human motion by optimizing together a SLAM camera trajectory and the human motion, utilizing motion priors to constrain the optimization.  
Differently, WHAM~\cite{shin2024wham} proposed training a regression network that given an input video and camera estimation directly predicts the global human trajectory. 
Recently, TRAM~\cite{wang2024tram} combined SLAM estimations with a metric depth network to further enhance the metric scale of camera trajectories. 
However, these methods are primarily designed for third-person, full-body motion, which presents challenges that are markedly different from those encountered in egocentric hand motion.
We propose a high-fidelity world-space 3D hand motion estimation approach to address these challenges effectively.


\section{Method}
\label{sec:method}
Given an egocentric video $\mathbf{V} \in\mathbb{R}^{T\times H\times W \times 3}$,  we aim to accurately reconstruct the complete 3D motion of hand $i$ represented with MANO~\cite{romero2017embodied} pose $\{\ti{\Theta} \in \mathbb{R}^{15 \times 3}\}_{t=0}^T$ and 
shape parameters $\{\beta^i_t \in \mathbb{R}^{10}\}_{t=0}^T$
along with a global orientation $\{\ti{\Phi} \in \mathbb{R}^3 \}_{t=0}^T$ and  
root translation $\{\ti{\Gamma} \in \mathbb{R}^3\}_{t=0}^T$ expressed in the world-coordinate system. The proposed method is composed of three main modules: i) the hand motion estimation network $\mathcal{M}$ that reconstructs robust hand motions in the camera-frame ii) the camera-trajectory estimation module that effectively predicts the camera pose in the world-coordinates and iii) the motion infiller network $\mathcal{F}$ that restores non-visible and occluded hands and reinforces the temporal coherence of the reconstructed 3D hand motion. 
An overview of the proposed framework is visualized in \cref{fig:method}.

\subsection{Hand Motion Estimation}

Predicting hand motion from egocentric videos presents significant challenges due to the prevalence of severe occlusions, motion blur and perspective distortions. 
Despite advancements in single-image hand pose estimation \cite{potamias2024wilor,hamer}, directly extending these methods to hand motion estimation presents three key challenges that limit accurate and robust reconstructions.
Firstly, such methods lack temporal coherence since they are trained on individual images, resulting in unpleasant jitter artifacts when applied to video reconstruction. 
Secondly, hands in egocentric videos often encounter a boundary truncation problem due to the limited field of view, leading to partial or incomplete hand visibility, which significantly deteriorates the performance of hand pose estimation frameworks.
Thirdly, the lack of motion priors in hand pose estimation methods, combined with severe occlusions and motion blur in egocentric videos, can further reduce the realism of reconstructed hand motions.
To effectively mitigate the aforementioned challenges, we propose a hand motion estimation network $\mathcal{M}$, which extends state-of-the-art single-image hand pose estimation methods \cite{potamias2024wilor} by learning spatio-temporal motion priors. 

In particular, given an input video $\mathbf{V}\in\mathbb{R}^{T\times H\times W \times 3}$, we first use multi-hand detection~\cite{potamias2024wilor} and tracking~\cite{aharon2022bot} methods to obtain the bounding box sequence of each hand $i$. We utilize the pre-trained ViT backbone of the state-of-the-art 3D hand reconstruction method WiLoR~\cite{potamias2024wilor} to extract robust image-aligned features $\mathbf{f}_t^{i}$ for each frame $t$ of the video. 
To mitigate truncated hands and thereby enhance the temporal consistency of the extracted image-aligned features, we introduce a temporal Image Attention Module (IAM) that updates the feature tokens $\mathbf{\hat{f}}_t^{i}$ with temporal information.
Using temporal self-attention, appearance features are fused across adjacent frames, enhancing the robustness of the backbone features at boundary regions. 
Following \cite{potamias2024wilor}, we utilize an additional token to regress MANO pose $\ti{\widetilde{\Theta}}$ and 
shape $\widetilde{\beta}^i_t$ parameters
along with the hand orientation $\widetilde{\Phi}^{c_t,i}_t$ and  
camera-space hand translation $\widetilde{\Gamma}^{c_t,i}_t$.

Nevertheless, despite IAM layer significantly enhancing the image features on truncated and occluded regions, the features extracted from ViT backbone still suffer from baked appearance and background elements and fail to capture expressive hand motion cues. 
To tackle this, we introduce an additional Pose Attention Module (PAM), which applies temporal self-attention directly to the MANO~\cite{romero2017embodied} pose parameters. 
Effectively, PAM learns hand motion priors to constrain the 3D reconstructions and improve the temporal coherence of motion.

\noindent{\textbf{Loss function.}} To train the hand motion estimation module $\mathcal{M}$, we utilize a set of loss functions, including 3D and 2D hand joint losses, along with direct MANO parameters supervision. The overall loss function is formulated as:
\begin{equation}
\begin{aligned}
    &\mathcal{L}_{\mathcal{M}} = \sum_{t=1}^{T}(\lambda_1 \mathcal{L}_{\text{3D}}^{t} + \lambda_2 
    \mathcal{L}_{\text{2D}}^{t} 
    + \lambda_3 \mathcal{L}_{\text{MANO}}^{t}), \\
    &\mathcal{L}_{\text{3D}}^{t} = ||\mathbf{J}_{\text{3D}}^{t} -\widetilde{\mathbf{J}}_{\text{3D}}^{t}||_1, \\
    &\mathcal{L}_{\text{2D}}^{t} = ||\mathbf{J}_{\text{2D}}^{t} -\widetilde{\mathbf{J}}_{\text{2D}}^{t}||_1, \\
    &\mathcal{L}_{\text{MANO}}^{t} = ||\Theta_{t} - \widetilde{\Theta}_{t}||_2^2 + ||\beta_{t} - \widetilde{\beta}_{t}||_2^2, 
\end{aligned}  
\end{equation}

where each $\lambda_i$ is a weighting factor that balances the influence of the respective loss terms.


\subsection{Camera Trajectory Estimation}
Estimating the camera motion in the world frame from an egocentric video can be viewed as a camera localization problem. 
However, despite the success of SLAM methods in addressing camera localization, two major challenges prevent their direct application to egocentric hand videos: 
Firstly, in egocentric videos, hands occupy a substantial portion of the field of view, which can highly influence the feature-matching step of structure-from-motion methods, leading to imprecise camera motions. 
Secondly, SLAM methods estimate camera translation up to an arbitrary scale, which does not reflect real-world translations. 
To tackle the aforementioned challenges, we propose a hybrid approach that leverages an adaptive SLAM method tailored to egocentric videos coupled with a foundational metric depth model to achieve robust camera pose estimation. 

\noindent{\textbf{Adaptive Egocentric SLAM.}} 
Despite advancements in SLAM methods, such as DROID-SLAM \cite{droid_slam}, which demonstrate robustness against subtle dynamic objects, large hand movements in egocentric views can severely impact the reconstruction accuracy of SLAM approaches.
Following \cite{wang2024tram}, we utilize a dual-masking strategy to exclude the hand motion from the reconstructed camera trajectory. 
In particular, we project the reconstructed 3D hand motions in the image space to define a hand mask $\mathbf{M}_t$.
We then mask the hand regions in both the input images and the predicted confidence maps of DROID-SLAM~\cite{droid_slam}. 
\begin{equation}
    \hat{I}_t = (1-\mathbf{M}_t) \cdot I_t,  \hat{w}_t = (1-\mathbf{M}_t) \cdot w_t
\end{equation}
This step eliminates the dynamic hand regions from both the feature extraction and the dense bundle adjustment steps of DROID-SLAM, making the camera trajectory estimation robust to dynamic hands. 
Specifically, masking the confidence map \(w_{t}\) effectively excludes the corresponding coordinates from the re-projection error calculation, ensuring that only background pixels contribute to camera motion estimation in the Dense Bundle Adjustment (DBA) process and enhances robustness against hand motion. 

\noindent{\textbf{Metric Scale Estimation.}} 
Given that monocular SLAM methods lack absolute depth information, they can only estimate the camera trajectory up to an arbitrary scale factor without a fixed world-scale reference.
Hence, SLAM methods can only estimate relative depths $\mathbf{d}_t$ in arbitrary units that do not correspond to a fixed scale.
To reconstruct high-fidelity camera translation scale $\alpha$ in real-world coordinates, we propose a robust scale estimation approach that integrates a metric network with a dynamic sampling. 
Specifically, we utilize Metric3D~\cite{metric3d}, a foundational model trained on large-scale datasets that can reliably predict metric-scale depth from a single image, ensuring generalization to in-the-wild data. 
For each keyframe of DROID-SLAM, we use Metric3D \cite{metric3d} to predict a scene depth $\mathbf{D}_t$ in meters. 
Furthermore, given that current metic networks are less accurate in regions too close and too far from the camera, we propose a dynamic sampling strategy to effectively increase the robustness of the scale estimation. 
Specifically, we mask out both the hand regions as well as points that are either near or far away from the camera and restrict the estimation of the scaling factor to reliable points within an intermediate range and outside the hand region.
The optimal min-max depth interval priors are derived by optimizing the scale accuracy on the egocentric training dataset.
Given the obtained hand masks and the thresholds for distance, the adaptive sampling module (AdaSM) selects a point set $S_t$ that satisfies:
\begin{equation}
    S_t = \{p \mid p \notin \mathbf{M}_t, \mathbf{D}_{\text{min}} < \mathbf{D}_t(p) < \mathbf{D}_{\text{max}}\}.
\end{equation}
Following \cite{wang2024tram}, we estimate the final scale $\alpha$ by optimizing the alignment between SLAM and Metric3D depth estimations on the sampled set as:
\begin{equation}
E(\alpha) = \sum_{p \in S_t} \mathcal{L}_{\text{GM}}({{\bf{D}}_{t}(p) - \alpha \cdot \bf{d}}_{t}(p) ),
\end{equation}
where $\mathcal{L}_{\text{GM}}$ is the German-McClure loss function~\cite{BarronCVPR2019}.
This approach optimizes the scale $\alpha$ estimation by focusing on regions where depth prediction is more reliable, thereby mitigating the influence of outliers such as moving hands or extreme depth values, achieving highly precise and robust scale estimation.

\subsection{Hand Motion Infiller}
Due to the limited field of view in egocentric videos, hands are often outside of the visible frame, leading to distorted and incomplete 3D reconstruction of the hand motion.
To address this issue, we introduce a novel hand motion infiller network $\mathcal{F}$, which is tailored to complete the out-of-bounds hands and reconstruct the full 3D hand motion sequence. 
In particular, given an incomplete $T$-frame sequence of hand $i$ MANO parameters of $\{\widetilde{\Theta}^i_t, \widetilde{\beta}^i_t, \widetilde{\Phi}^{c_t,i}_t, \widetilde{\Gamma}^{c_t,i}_t \}$ predicted from the hand motion estimation network in each camera frame $c_t$ (with missing frames set to zero), the motion infiller network predicts a complete motion $\{\widehat{\Phi}^i_t, \widehat{\Theta}^i_t, \widehat{\beta}^i_t, \widehat{\Gamma}^i_t \}$ that accurately fills the missing frames. 


\noindent{\textbf{Canonical space transformation.} 
As a first step, we transform the input sequence from camera space to canonical space, which decouples the hand motion from the dynamic camera and aligns the sequence start state to zero translation and zero rotation. 
This operation can standardize the input sequence and facilitate training. Specifically, we first compute the camera-to-canonical transformation $T^{c_{i}2cano,i}=[R^{c_{i}2cano,i}|t^{c_{i}2cano,i}]$ that aligns the first frame's hand rotation and translation to zero. 
Subsequently, the hand rotations and translations are transformed into canonical space:
\begin{equation}
\begin{aligned}
    &\Phi^{cano,i}_t=R^{c_{t}2cano,i} \times \Phi^{c_t,i}_t, \\
    &\Gamma^{cano,i}_t=R^{c_{t}2cano,i} \times \Gamma^{c_t,i}_t + t^{c_{t}2cano,i}. \\ 
\end{aligned}
\end{equation}

\noindent \textbf{Infiller Network.} Predicting the hand pose of missing frames can be considered a motion-in-between task. 
To this end, we follow \cite{kim2022conditional} and build our motion infiller network using a transformer-encoder architecture trained to predict the missing pose tokens. 
Specifically, we initially project the input MANO sequences to $D$-dimension latent vectors and then feed them to a set of stacked multi-head self-attention layers.
Given that transformer encoder does not explicitly capture the auto-regressive nature of the motion, we incorporate positional embeddings~\cite{vaswani2017attention} to encode the temporal information of each frame. 
The output tokens are passed to a simple fully-connected decoder that regresses the MANO sequence in canonical space. 
Finally, we convert the MANO sequence to the world space by computing the canonical-to-world transformation $T^{cano2w,i}=[R^{cano2w,i}|t^{cano2w,i}]$.

\noindent{\textbf{Training.}} 
To train the motion infiller network, we use HOT3D~\cite{banerjee2024introducing} since it provides both egocentric and third-person views of the hands, enabling us to easily identify and label the frames of each video where the hands are out of the egocentric camera frustum. 
To augment the training data, we sample additional video sequences and randomly mask frame segments while retaining the start and the end frames to serve as context for the infiller network.
To facilitate the training process, we initialize the MANO parameters of the missing frames using a pose interpolation scheme. 
Specifically, for a given motion sequence, translations and shape parameters are linearly interpolated, while global rotations and pose parameters are interpolated with spherical linear interpolation (SLERP).
This can reduce the workload of the infiller network and enable more robust reconstructions. 

\noindent{\textbf{Loss Functions.}} We train the motion infiller network using a combination of loss functions to penalize the world translation and orientation along with the hand pose and shape. 
The overall loss function is formulated as:
\begin{equation}
\begin{aligned}
    &\mathcal{L}_{\mathcal{F}} = \sum_{t=1}^{T} (\gamma_1 \mathcal{L}_{\Gamma}^t + \gamma_2 \mathcal{L}_{\Phi}^t + \gamma_3 \mathcal{L}_{\Theta}^t + \gamma_4 \mathcal{L}_{\beta}^t),  \\
    &\mathcal{L}_{\Gamma}^t = ||\Gamma_t -\hat{\Gamma}_t||_1, 
    \mathcal{L}_{\Phi}^t = ||\Phi_t -\hat{\Phi}_t||_1, \\
    &\mathcal{L}_{\Theta}^t = ||\Theta_t -\hat{\Theta}_t||_1, 
    \mathcal{L}_{\beta}^t = ||\beta_t - \hat{\beta}_t||_1,
\end{aligned}
\end{equation}
where each $\gamma_i$ is a weighting factor that balances the influence of the respective loss terms.

\section{Experiments}
\label{sec:exp}
\noindent\textbf{Datasets.} To assess the camera-frame hand motion reconstruction performance of HaWoR and baseline models we utilize DexYCB~\cite{chao2021dexycb} dataset, which comprises videos capturing hand-object interactions from a set of static cameras including conditions of severe occlusion. 
To evaluate the reconstructed world-space camera and hand trajectories along with the infiller reconstructions, we use HOT3D dataset~\cite{banerjee2024introducing}, that contains egocentric videos captured from dynamic cameras accompanied with ground-truth camera trajectories along with MANO annotations in world-coordinates. 



\noindent\textbf{Evaluation Metrics.}  
To evaluate the 3D hand pose in the camera-frame, we use Procrustes-Aligned Mean Per Joint Position Error (PA-MPJPE) and the Area Under the Curve (AUC) of correctly localized keypoints. Following~\cite{slahmr}, we assess the hand estimation in the world-frame using World MPJPE (W-MPJPE) and World Aligned MPJPE (WA-MPJPE). In addition, we evaluate the error of the entire trajectory with root translation error (RTE) and compute acceleration error (Accel) to evaluate the smoothness of motion. Frechet Inception Distance (FID) is used to measure the motion filling quality. To quantify the quality of the camera trajectory, we compute the Average Trajectory Error (ATE) that aligns the scale of GT and ATS-S that uses the estimated scale, as described in \cite{wang2024tram}.

\subsection{Camera-frame 3D Hand Motion}
To achieve accurate world-space hand motion reconstruction it is essential to achieve robust and high fidelity hand motion estimation in the camera-frame. 
Given that egocentric videos often face sever occlusions, we follow~\cite{fu2023deformer} and utilize DexYCB dataset that provides explicit annotations regarding the occluded frames within a video. 
Specifically, in~\cref{tab:res_dexycb} we compare HaWoR against image- and video-based methods for camera-frame 3D hand motion reconstruction under different occlusion ratio levels. 
As can be observed, HaWoR archives robust performance across different occlusion rates. 
In contrast, WiLoR~\cite{potamias2024wilor} and Deformer~\cite{fu2023deformer} that serve as state-of-the-art methods for 3D hand pose estimation from single-image and video, respectively, face a huge performance degradation on videos with increased occlusion rates.
It is important to note that HaWoR performance on sever occlusion rate (75\%-100\%) shows a more significant improvement than state-of-the-art methods (\ie, WiLoR~\cite{potamias2024wilor} 5.68 vs HaWoR 5.07).
\begin{table}[t]
\centering
\small{
\setlength{\tabcolsep}{0.5pt}
\begin{tabular}{clcccccc}
\toprule 
 \multirow{2}{*}{} & \multirow{2}{*}{Methods} & \multicolumn{2}{c}{All} & \multicolumn{2}{c}{50\%-75\%} & \multicolumn{2}{c}{75\%-100\%} \\
 \cmidrule(lr){3-4}  \cmidrule(lr){5-6}  \cmidrule(lr){7-8}
& & MPJPE & AUC & MPJPE & AUC & MPJPE & AUC \\
\midrule  
\multirow{6}{1em}{\rotatebox[origin=c]{90}{mococular}}  & Spurr \etal \cite{spurr2020weakly}  & 6.83 & 86.4 & 8.00 & 84.0 & 10.65 & 78.8\\
& MeshGraphormer~\cite{lin2021mesh}  & 6.41 & 87.2 & 7.22 & 85.6 & 7.76 & 84.5 \\
& SemiHandObj \cite{liu2021semi}  & 6.33 & 87.4 & 7.17 & 85.7 & 8.96 & 82.1 \\
& HandOccNet \cite{park2022handoccnet}  & 5.80 & 88.4 & 6.43 & 87.2 & 7.37 & 85.3 \\
& {WiLoR}~\cite{potamias2024wilor} & \cellcolor{TableLightGreen}5.01 & \cellcolor{TableLightGreen}90.0 & \cellcolor{TableLightGreen}5.42 & \cellcolor{TableLightGreen}89.2 &\cellcolor{TableLightGreen} 5.68 &\cellcolor{TableLightGreen} 88.7\\
\midrule
\multirow{5}{1em}{\rotatebox[origin=c]{90}{temporal}} & $S^2$HAND(V)~\cite{tu2023consistent} &  7.27 & 85.5 & 7.71 & 84.6 & 7.87 & 84.3 \\
& VIBE~\cite{kocabas2020vibe} & 6.43 & 87.1 & 6.84 & 86.4 & 7.06 & 85.8 \\
& TCMR~\cite{choi2021beyond} & 6.28 & 87.5 & 6.58 & 86.8 & 6.95 & 86.1 \\
& Deformer~\cite{fu2023deformer} & \cellcolor{TableYellow}5.22 & \cellcolor{TableYellow}89.6 & \cellcolor{TableYellow}5.70 & \cellcolor{TableYellow}88.6 & \cellcolor{TableYellow}6.34 & \cellcolor{TableYellow}87.3 \\ 
\cmidrule{2-8}
& \textbf{Proposed} & ~{\textbf{\cellcolor{TableDarkGreen}4.76}} & ~{\textbf{\cellcolor{TableDarkGreen}90.5}} & ~{\textbf{\cellcolor{TableDarkGreen}5.03}} & ~{\textbf{\cellcolor{TableDarkGreen}89.9}} & ~{\textbf{\cellcolor{TableDarkGreen}5.07}} & ~{\textbf{\cellcolor{TableDarkGreen}89.9}} \\
\bottomrule
\end{tabular}
\caption{Quantitative camera-frame comparison of state-of-the-art hand pose estimation methods on the \textbf{DexYCB} test dataset. We compare PA-MPJPE and AUC results, especially the split under large occlusion proportion (50\%-75\% and 75\%-100\%), which highlights our robustness in challenging visibility conditions.}
\label{tab:res_dexycb}}
\end{table}

\begin{figure*}[t]
  \centering
  \includegraphics[width=1\textwidth]{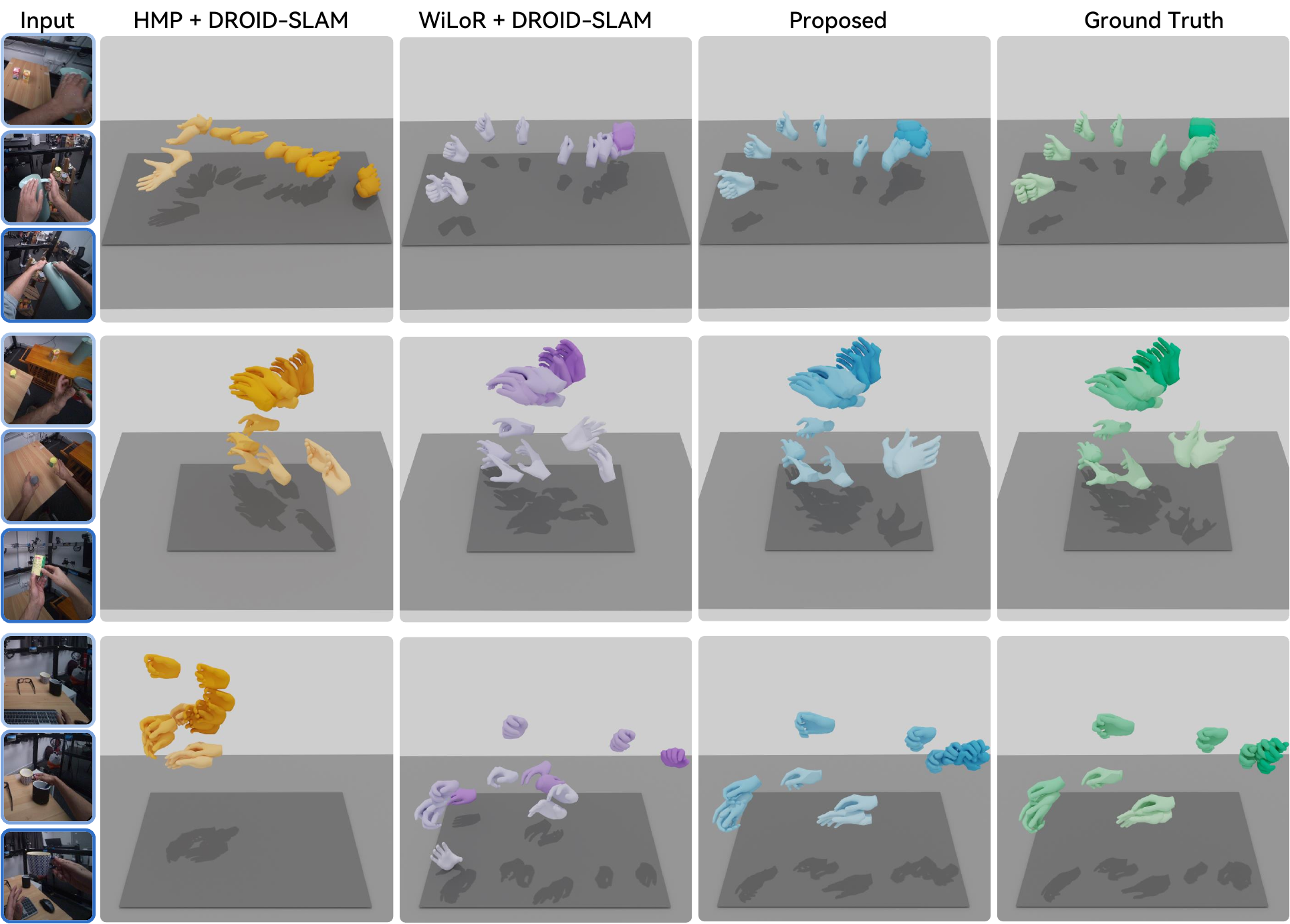}
  \caption{\textbf{Visualization} of right-hand estimated trajectories on challenging cases of \textbf{HOT3D}. The first example depicts \textit{someone picking up a kettle, turning around, and pouring water}. The second example depicts \textit{the subject placing a tin on the table and then picking up another}. The third video depicts \textit{the subject using a mouse keyboard and then reaching for a cup to drink water}.
  In contrast to the baseline methods, HaWoR achieves robust hand trajectories, especially in challenging scenarios with large hand movements and truncated hands.
  }
  \label{fig:hand_vis}
\end{figure*}

\subsection{World-frame 3D Hand Motion}
In this section, we quantitatively and qualitatively evaluate HaWoR in hand motion reconstruction in the world-space. 

\noindent\textbf{Baselines.} Given that HaWoR is currently the first, to the best of our knowledge, framework that tackles world-space hand motion reconstruction, we implement a set of strong baseline methods that follow the literature of world-grounded human body motion estimation~\cite{glamr,slahmr,shin2024wham}.
In particular, we use state-of-the-art performing methods for hand pose estimation, namely HaMeR~\cite{hamer}, WiLoR~\cite{potamias2024wilor} and HandDGP~\cite{valassakis2025handdgp}, coupled with DROID-SLAM~\cite{droid_slam}, to recover the world-space hand and camera motion. 
Additionally, we implement an optimization-based method that closely follows SLAHMR~\cite{slahmr} by combining DROID-SLAM with the powerful hand motion prior (HMP)~\cite{hmp} to align the hand pose estimations with the camera trajectories. 
Additionally, we compare the proposed world-frame camera trajectory with different metric depth estimation methods, including ZoeDepth~\cite{zoedepth}, DepthAnythingV2~\cite{depthanything} and Metric3DV2~\cite{metric3d}. 
To facilitate understanding between the contribution of each network component, we divide the evaluation into two steps to assess both the global camera trajectory and the reconstructed hand motion in world-coordinates. 

\begin{table}[t!]
\centering
\setlength{\tabcolsep}{1pt}
\small
\begin{tabular}{lccccc}
\cmidrule{1-6}
\multirow{2}{*}{Methods}  & ATE$\downarrow$ & \multicolumn{4}{c}{ATE-S$\downarrow$} \\
\cmidrule{2-2} \cmidrule(lr){3-6}
& All &  Short & Med & Long & All \\
\cmidrule{1-6} 
DROID~\cite{droid_slam} & 3.80 & - & - & - & - \\
DROID + ZoeDepth~\cite{zoedepth} & 3.80 & 25.03 & 39.39 & 75.95 & 43.58 \\
DROID + DepthAnyV2~\cite{depthanything} & 3.80 & 18.14 & 25.50 & 43.60 & 27.49  \\
DROID + Metric3DV2~\cite{metric3d} & 3.80 & 14.28 & \cellcolor{TableYellow}21.56 & 29.10 & 21.07\\
\cmidrule{1-6}
Proposed w/o Scale & 3.36 & - & - & - & - \\ 
Proposed w. ZoeDepth~\cite{zoedepth} & 3.36 & \cellcolor{TableLightGreen}11.91 & 25.34 & 36.05 & 23.67\\
Proposed w. DepthAnyV2~\cite{depthanything}  & 3.36 & 14.63 & \cellcolor{TableLightGreen}20.49 & \cellcolor{TableLightGreen}25.54 & \cellcolor{TableLightGreen}19.85 \\
Proposed w/o AdaSM & 3.36 & \cellcolor{TableYellow}14.03 & 22.38 & \cellcolor{TableYellow}27.49 & \cellcolor{TableYellow}20.97 \\
\midrule
\textbf{Proposed} & ~{\textbf{\cellcolor{TableDarkGreen}3.36}} &  ~{\textbf{\cellcolor{TableDarkGreen}9.31}} & ~{\textbf{\cellcolor{TableDarkGreen}15.86}} & ~{\textbf{\cellcolor{TableDarkGreen}19.26}} & ~{\textbf{\cellcolor{TableDarkGreen}14.61}}  \\
\cmidrule{1-6}
\end{tabular}
\caption{
Evaluation of camera estimation with aligned scale (\textbf{ATE}) and estimated scale \textbf{(ATE-S)}. We also report the split results of short ($<5m$), medium ($3m-5m$) and long ($>5m$) displacement. ATE and ATE-S is in $mm$. 
}
\label{tab:eval_hot3d_cam}

\end{table}

\begin{table}[tb]
  \centering
  \setlength{\tabcolsep}{1pt}
  \small
  \begin{tabular}{lccccc}
    \toprule
       \footnotesize Method & \footnotesize PA-MPJPE & \footnotesize W-MPJPE & \footnotesize WA-MPJPE & \footnotesize RTE & \footnotesize Accel\\
    \midrule
        HaMeR-SLAM~\cite{hamer} & \cellcolor{TableYellow}9.39 & 156.03 & 43.37 & 4.77 & 19.25 \\
        HandDGP-SLAM~\cite{valassakis2025handdgp} & 17.88 & 154.30 & 42.93 & 3.18 &  20.17 \\
        WiLoR-SLAM~\cite{potamias2024wilor} & \cellcolor{TableLightGreen}6.00 & \cellcolor{TableYellow}151.67 & \cellcolor{TableYellow}39.49 & \cellcolor{TableYellow}2.99 & \cellcolor{TableYellow}8.02 \\

        HMP-SLAM~\cite{hmp} & 10.51 & \cellcolor{TableLightGreen}119.41 & \cellcolor{TableLightGreen}39.46 & \cellcolor{TableLightGreen}2.79 & \cellcolor{TableLightGreen}5.50 \\
    \midrule   
        \textbf{Proposed}            & ~{\textbf{\cellcolor{TableDarkGreen}4.79}}  & ~{\textbf{\cellcolor{TableDarkGreen}33.20}} & ~{\textbf{\cellcolor{TableDarkGreen}11.27}} & ~{\textbf{\cellcolor{TableDarkGreen}0.78}} & ~{\textbf{\cellcolor{TableDarkGreen}5.41}}  \\
  \bottomrule
  \end{tabular}
  \caption{Quantitative evaluation in world-space coordinates on \textbf{HOT3D} dataset. 
  }
  \label{tab:world_space_hot3d}
\end{table}

\noindent\textbf{Global Camera Trajectory.}
In \cref{tab:eval_hot3d_cam} we evaluate the camera trajectory estimation of the proposed adaptive egocentric SLAM method compared to baseline approaches that naively combine DROID-SLAM~\cite{droid_slam} with metric networks. 

As can be easily observed, although directly applying metric models to SLAM-estimated camera trajectories may be sufficient for third-person body motion reconstruction~\cite{glamr,slahmr,shin2024wham,yin2024whac}, it falls short in accurately reconstructing camera trajectories in egocentric scenarios, which are characterized by significant occlusions and hands occupying a large portion of the frame.
In contrast, the proposed approach effectively mitigates this issue by providing accurate visual cues during the SLAM bundle adjustment step that facilitate the reconstruction performance in egocentric scenarios. 
It is also important to note that the trajectory error is further exacerbated when it comes to estimating the actual world-scale estimations (ATE-S). 
\begin{figure}[t]
  \centering
  \includegraphics[width=1\columnwidth]{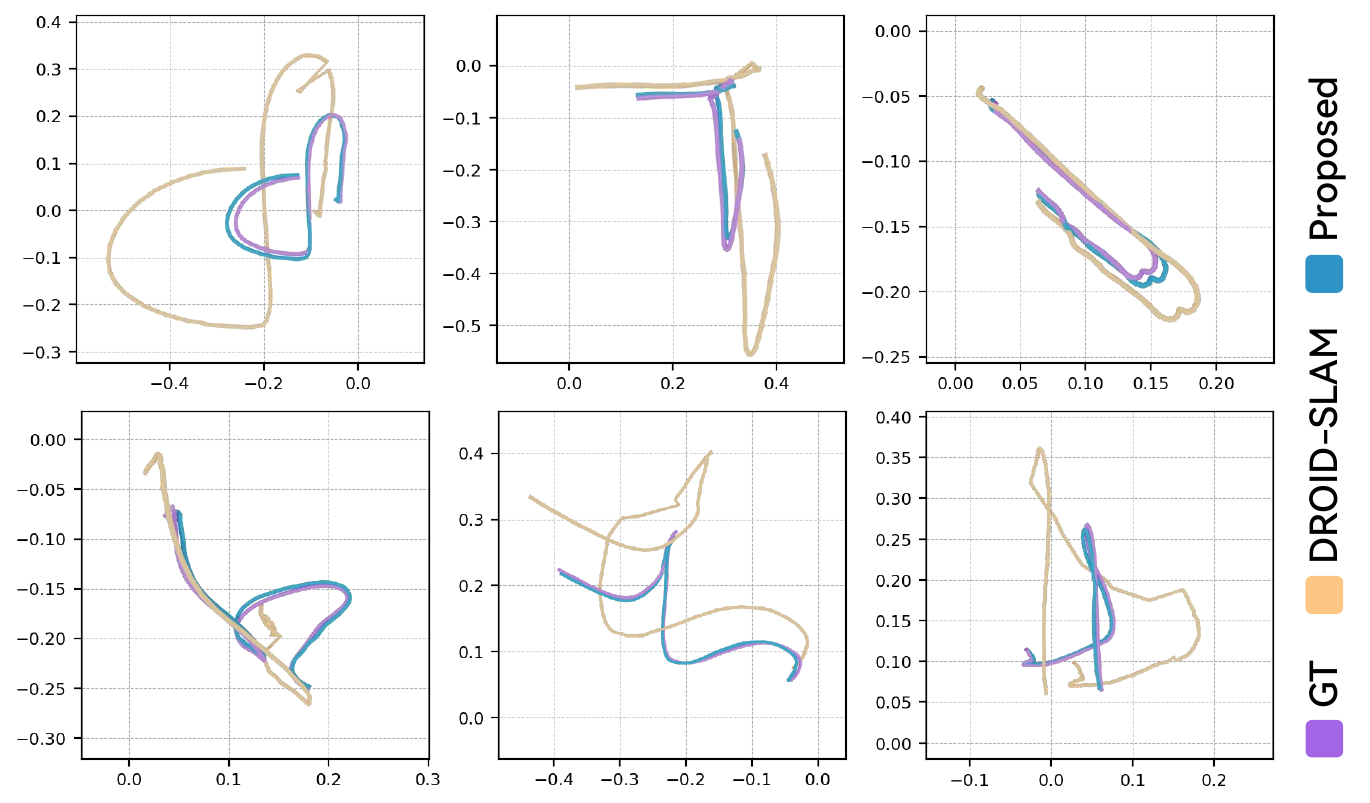}
  \caption{\textbf{Camera global trajectory}. The proposed adaptive SLAM approach demonstrates precise camera trajectory estimation while recovering accurate real-world scale, outperforming DROID-SLAM, which struggles with both trajectory accuracy and scale consistency.}
  \label{fig:cam_traj}
\end{figure}
The effect of the adaptive SLAM proposed in HaWoR can be further validated in \cref{fig:cam_traj}, where the camera trajectories match the ground truth camera motion, addressing the limitations of DROID-SLAM approach in egocentric views. 

\noindent\textbf{Hand motion estimation in world-coordinates.}
In \cref{tab:world_space_hot3d} we report the performance of HaWoR and the baseline methods in hand motion reconstruction in the world-coordinates. 
HaWoR significantly outperforms both regression and optimization-based baselines by a large margin under both camera (PA-MPJPE) and world-space reconstructions (W-MPJPE). 
Furthermore, HaWoR produces more stable and robust motion reconstructions that are unaffected by occlusions, as indicated by the RTE metric. Besides, compared to the baseline methods, HaWoR achieves significantly lower acceleration error, validating the smoothness of the reconstructed motions across frames.

The accuracy of the proposed hand motion can be further validated in both  ~\cref{fig:hand_vis} and ~\cref{fig:hand_traj}, where we compare the hand trajectories estimated from HaWoR and an optimization-based approach that utilizes hand motion priors to guide the motion (\textit{HMP-SLAM}). 
HaWoR achieves accurate hand trajectories that follow the ground truth even on complex motions that the baseline methods fail. 
It is also important to note that apart from the superior performance in camera trajectory estimation, HaWoR requires only a single forward pass of 40 $ms$ per frame, significantly reducing inference runtime by 75\% compared to optimization-based method of HMP-SLAM, that requires 160 $ms$ for per frame.
\begin{figure}[t]
  \centering
  \includegraphics[width=1\columnwidth]{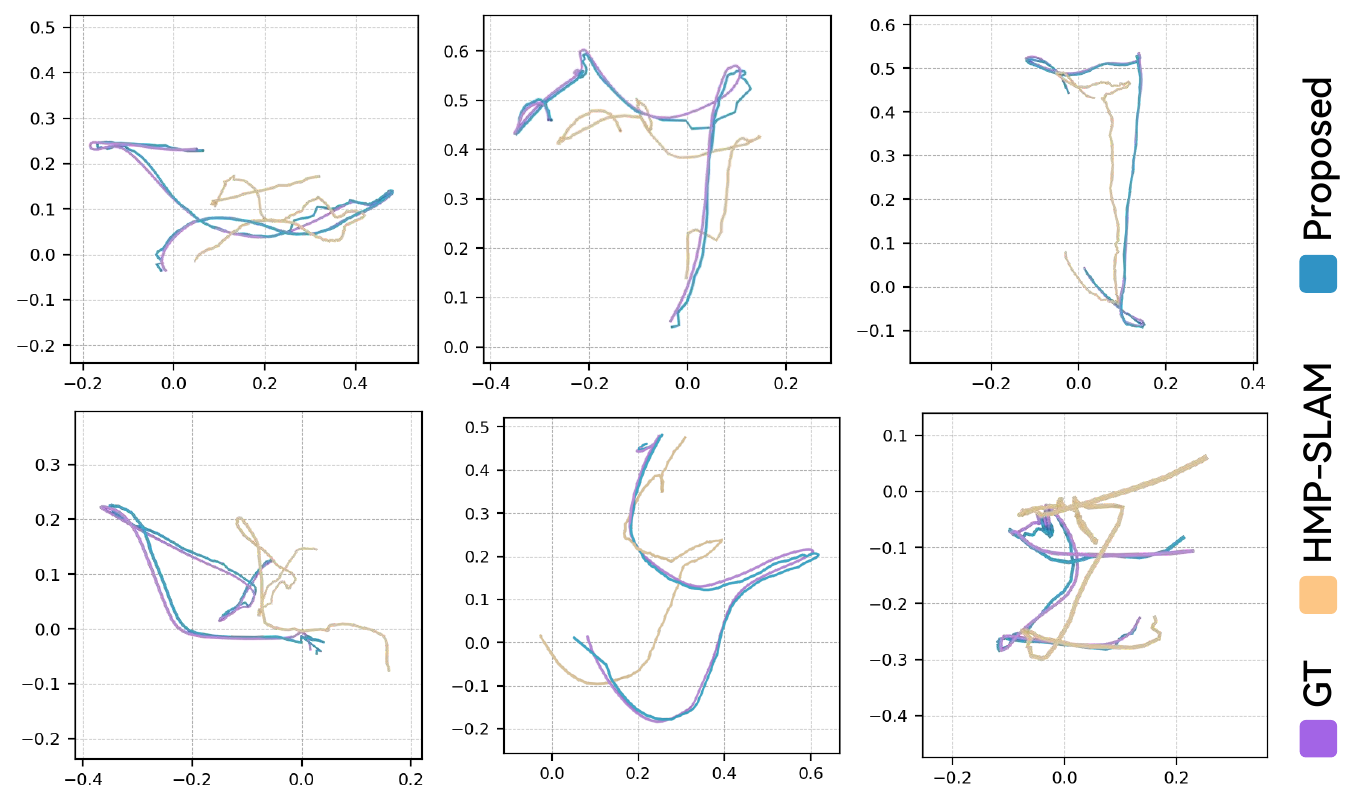}
  \caption{\textbf{Hand global trajectory} for the right hand on HOT3D. Compared to HMP-SLAM, HaWoR produces accurate trajectories even for complex and long-range hand movements.}
  \label{fig:hand_traj}
\end{figure}



\begin{table}[]
\small
\centering
\begin{subtable}[t]{1\columnwidth}
\centering
\setlength{\tabcolsep}{1pt}
\begin{tabular}{lccccc}
\toprule
\footnotesize Method & \footnotesize PA-MPJPE & \footnotesize W-MPJPE & \footnotesize WA-MPJPE& \footnotesize RTE & \footnotesize Accel\\ 
\midrule
w/o Pretrained ViT & 7.59 & 86.80 & 19.46 & 1.26 & 9.09\\
w/o IAM \& PAM & 5.07 & 44.60 & 13.85 & 0.93 & 8.42 \\
w/o PAM & 4.80 & 36.32 & 12.40 & 0.88 & 6.03 \\
\midrule
\textbf{Proposed} & \textbf{4.79} & \textbf{33.20} & \textbf{11.27} & \textbf{0.78} & \textbf{5.41} \\
\bottomrule
\end{tabular}
\caption{\textbf{Hand motion components}. 
Here is the ablation results without IAM (Image Attention Module), PAM (Pose Attention Module) or pretrained ViT. }
\end{subtable}
\hfill



\begin{subtable}[t]{1\columnwidth}
\centering
\setlength{\tabcolsep}{2pt}
\begin{tabular}{lcccccc}
\toprule
Method & FID & PA-MPJPE &  W-MPJPE &  WA-MPJPE & RTE \\ 
\midrule
Last Pose & 1.52 & 7.83& 116.79 & 78.78 & 13.04\\
LERP & 1.42 & 6.33 & 75.01 & 49.16 & 9.39\\
\midrule
\textbf{Proposed} & \textbf{0.57} & \textbf{6.22} & \textbf{66.25} & \textbf{37.22} & \textbf{7.41}\\
\bottomrule
\end{tabular}
\caption{\textbf{Motion Infiller}. We experiment on the invisible sequences of HOT3D~\cite{banerjee2024introducing} validation dataset.}
\end{subtable}
\caption{Ablations study on the key modules of our method.}
\label{tab:ablation}
\end{table}





\subsection{Ablation}
We perform an ablation study to assess the effect of key components in our framework.
In particular, we initially report the effect of the image and pose motion priors that compose the proposed hand motion estimation network. 
As can be seen in \cref{tab:ablation}(a), both IAM and PAM modules contribute to the performance of HaWoR, improving the robustness of the reconstructions. 
Furthermore, we evaluate the contribution of the motion infiller network and its generalization performance on HOT3D~\cite{banerjee2024introducing} datasets. 
LERP is using frame linear interpolation, where root translation and shape are linearly interpolated, and joint rotations are spherically linear interpolated. We also compare with replicating the last visible pose.
As can be observed from \cref{tab:ablation}(b), the proposed motion infiller network can significantly outperform naive motion completion methods.  

\section{Conclusion and Limitations}
In this work we present HaWoR, a high-fidelity 3D hand motion reconstruction method in the world-space. 
HaWoR is founded on a powerful camera-frame transformer-based hand motion reconstruction module and a robust infiller network to estimate and fill the motion-in-between missing frames. 
To align the camera-frame hand motions in the world-coordinate system we propose an adaptive egocentric SLAM module that facilitates global camera trajectory estimation under challenging and occluded egocentric views. 
Through extensive experimental results we demonstrate that HaWoR outperforms previous methods and achieves state-of-the-art performance under different benchmark datasets. 
However, while HaWoR significantly accelerates hand motion reconstruction compared to previous approaches, the runtime performance is still far from real-time. 
In the future, we could explore foundational models to directly estimate world-space camera trajectories to make a step towards real-time world-frame hand motion estimation. 


{
    \small
    \bibliographystyle{ieeenat_fullname}
    \bibliography{main}
}

\clearpage
\setcounter{page}{1}
\maketitlesupplementary

\begin{figure*}[hb]
  \centering
  \includegraphics[width=1\textwidth]{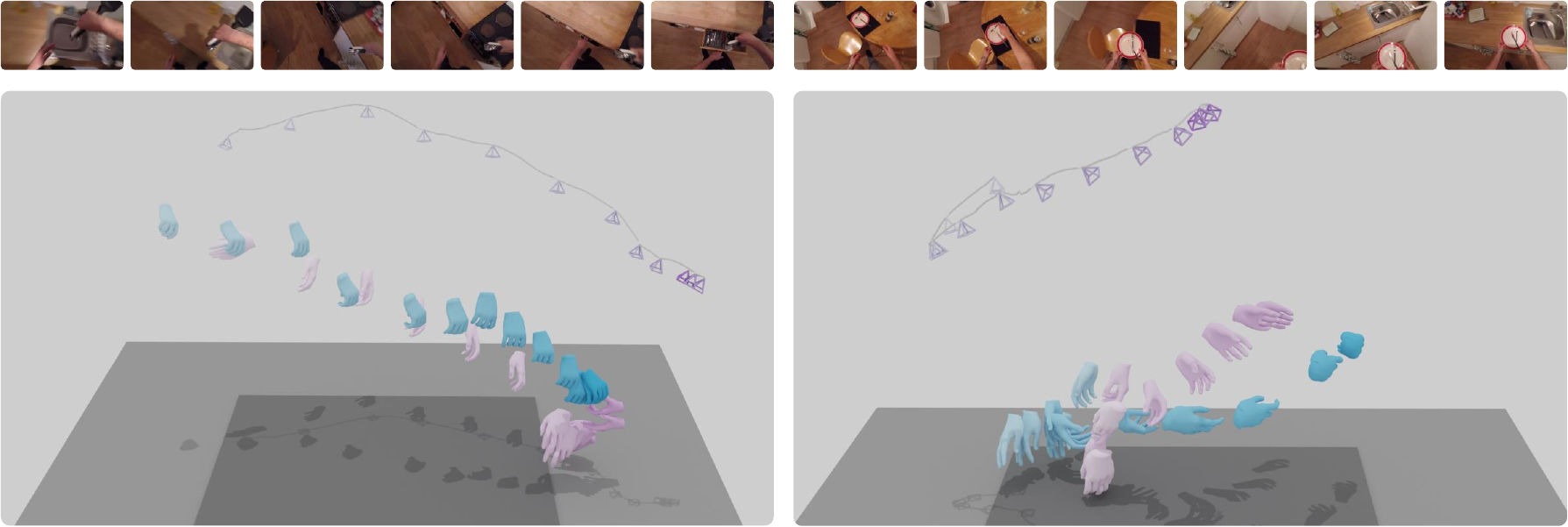}
  \caption{\textbf{Qualitative Evaluation } of the reconstructed world-space hands on in-the-wild videos from EPIC-KITCHENS~\cite{damen2018scaling}. Refer to the supplementary video.
  }
  \label{fig:itw}
\end{figure*}

\section{Generalization on In-the-Wild Videos}

To evaluate the generalization of HaWoR on in-the-wild video, we show qualitative results of the camera and hand reconstruction on sequences from EPIC-KITCHENS~\cite{damen2018scaling} in ~\cref{fig:itw}. 
Although the proposed model has not been trained on these in-the-wild data, it can still recover hands and cameras that are consistent with the input videos. 
We include additional in-the-wild cases in the supplementary video, where the generalization of HaWoR can be easily observed.
It is worth noting that the input videos include numerous frames where the hands are outside the view frustum. Despite this, HaWoR achieves accurate reconstructions by leveraging the proposed motion priors and the infilling network.

\begin{figure}[htb]
  \centering
  \includegraphics[width=1\columnwidth]{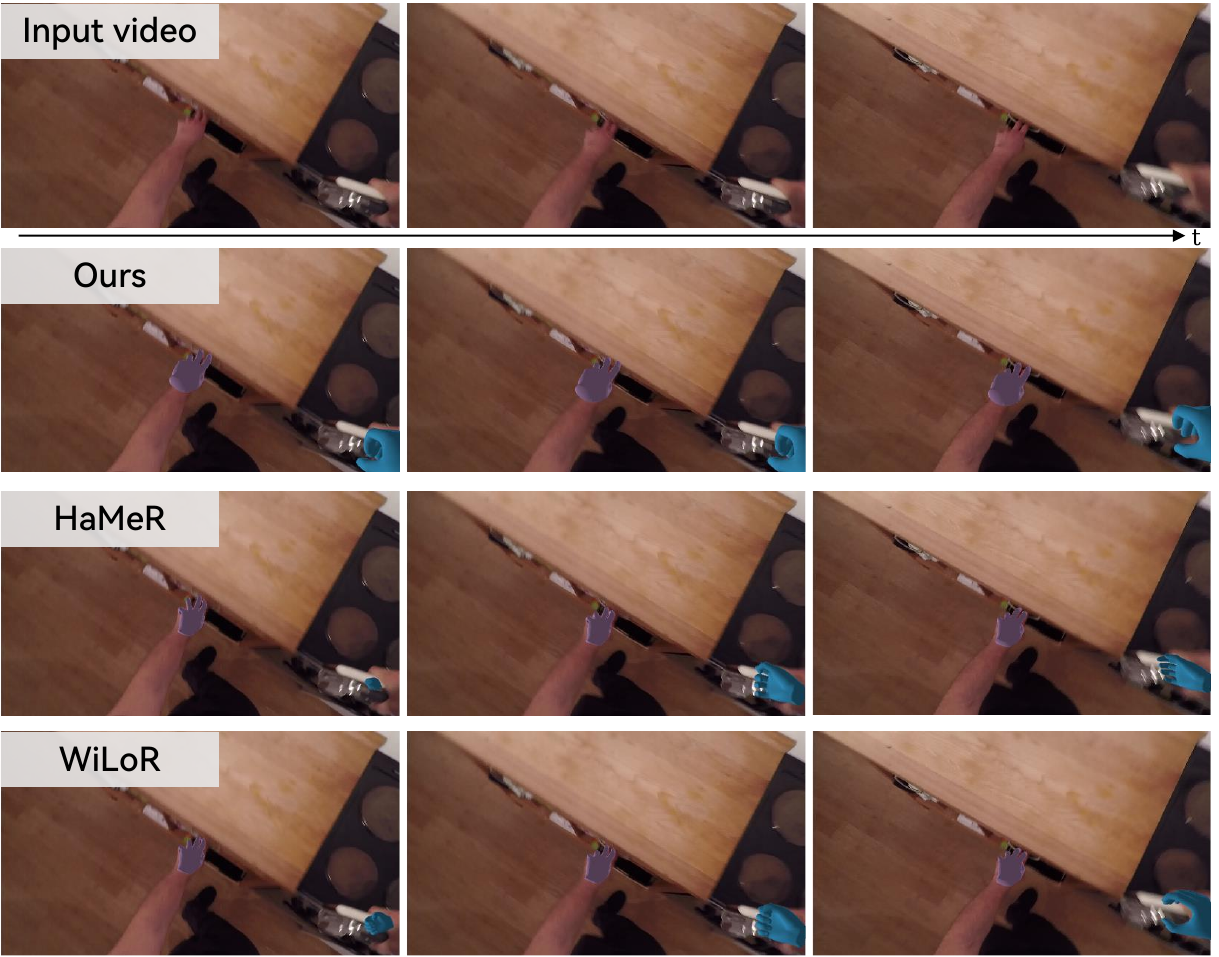}
  \caption{\textbf{Qualitative Comparison} with other state-of-the-art methods on in-the-wild videos from EPIC-KITCHENS~\cite{damen2018scaling}. Refer to the supplementary video.
  }
  \label{fig:comparison}
\end{figure}

\section{Comparison on In-the-Wild Videos}

To compare HaWoR with other state-of-the-art methods on in-the-wild video, we show qualitative results in the camera view on sequences from EPIC-KITCHENS~\cite{damen2018scaling} in ~\cref{fig:comparison}. It is evident that our method achieves significantly better results compared to HaMeR~\cite{hamer} and WiLoR~\cite{potamias2024wilor} when hand truncation occurs at the boundary.

\section{Implementation Details}

In this section we provide the training and evaluation details about the network of hand motion estimation and hand motion infiller. 

\subsection{Hand Motion Estimation Network}

To train the hand motion estimation network we use a combination of multiple hand video datasets for generalization of the model. In particular, we use 4 video datasets with both 3D and 2D hand annotations, constructing of 1M training frames totally:

\begin{itemize}
    \item HOT3D~\cite{banerjee2024introducing} is an egocentric video dataset that contains daily hand activity, and we partitioned it to use 573K frames as training set. 
    \item ARCTIC~\cite{fan2023arctic} dataset that contains two hands dexterously manipulating objects, focusing on hand-object interaction dynamics and 165K video frames are utilized for training. 
    \item DexYCB~\cite{chao2021dexycb} is a dataset focused on capturing hand grasping of objects, designed to support tasks in hand-object interaction and robotics, which provides 169K frames to train.
    \item HO3D~\cite{hampali2020honnotate} is a markerless dataset of color images with hands and objects involving 10 persons and 10 objects, and there are 66K frames for training. 
\end{itemize}

\noindent We train the hand motion estimation network with AdamW~\cite{loshchilov2017decoupled} optimizer for 250K iterations with a learning rate of 1e-5. The model is trained by freezing the pretrained ViT backbone of WiLoR~\cite{potamias2024wilor} for 2 days, using four NVIDIA A800 and a total batch size of 32. Regarding the loss weighting factor, we set $\lambda_1=0.05$ for the 3D keypoint loss, $\lambda_2=0.01$ for the 2D keypoint loss and $\lambda_3=0.001$ for the MANO pose loss. 

\subsection{Hand Motion Infiller}

To facilitate the training, we transform the input sequence from camera space to canonical space, as shown in ~\cref{fig:camera2cano}. Specifically, we define the first frame of each sequence as the canonical frame, and we compute the canonical transformation by aligning the first ($0^{th}$) frame’s hand rotation and offset the hand translation to zero:
{\small
\begin{equation}
\begin{aligned}
    & R^{c_t2cano,i} = (R_{c_0} \times \Phi^{c_0,i}_{0})^{-1} \times R_{c_t}, \\
    & t^{c_t2cano,i} = (R_{c_0} \times \Phi^{c_0,i}_{0})^{-1}(t_{c_t}-t_{c_0}-R_{c_0} \times \Gamma_{0}^{c_0,i}),
\end{aligned}
\end{equation}
}

where $R_{c_t}$ denotes the rotation of $t^{th}$ frame camera to world, $t_{c_t}$ is the translation of $t^{th}$ frame camera to world, $\Phi^{c_0,i}_{0}$ and $\Gamma^{c_0,i}_{0}$ are the hand rotation and translation in $0^{th}$ frame camera space.

To train the hand motion infiller we use the HOT3D~\cite{banerjee2024introducing} dataset that provides 3D hand annotations of all frames, including the frames with missing hands. We first collect the non-visible hand segments to create training sequences. To increase the data scale, other sequences are sampled from the dataset and randomly masked. We keep the start and end frames as context for the infiller and randomly mask middle continuous frames. We train the hand motion infiller with AdamW~\cite{loshchilov2017decoupled} optimizer for 1500K iterations. The learning rate is initialized with 0.0001 and decreased by a factor of 0.9 every 100 steps. We trained the model for 1 day using one NVIDIA A800 and a batch size of 32. For weighting the losses, we set $\gamma_1=0.05$ for the translation loss, $\gamma_2=2.0$ for the rotation loss, $\gamma_3=2.0$ for the pose loss and $\gamma_4=0.05$ for the shape loss.


\subsection{Evaluation Details}

\begin{figure}[t]
  \centering
  \includegraphics[width=1\columnwidth]{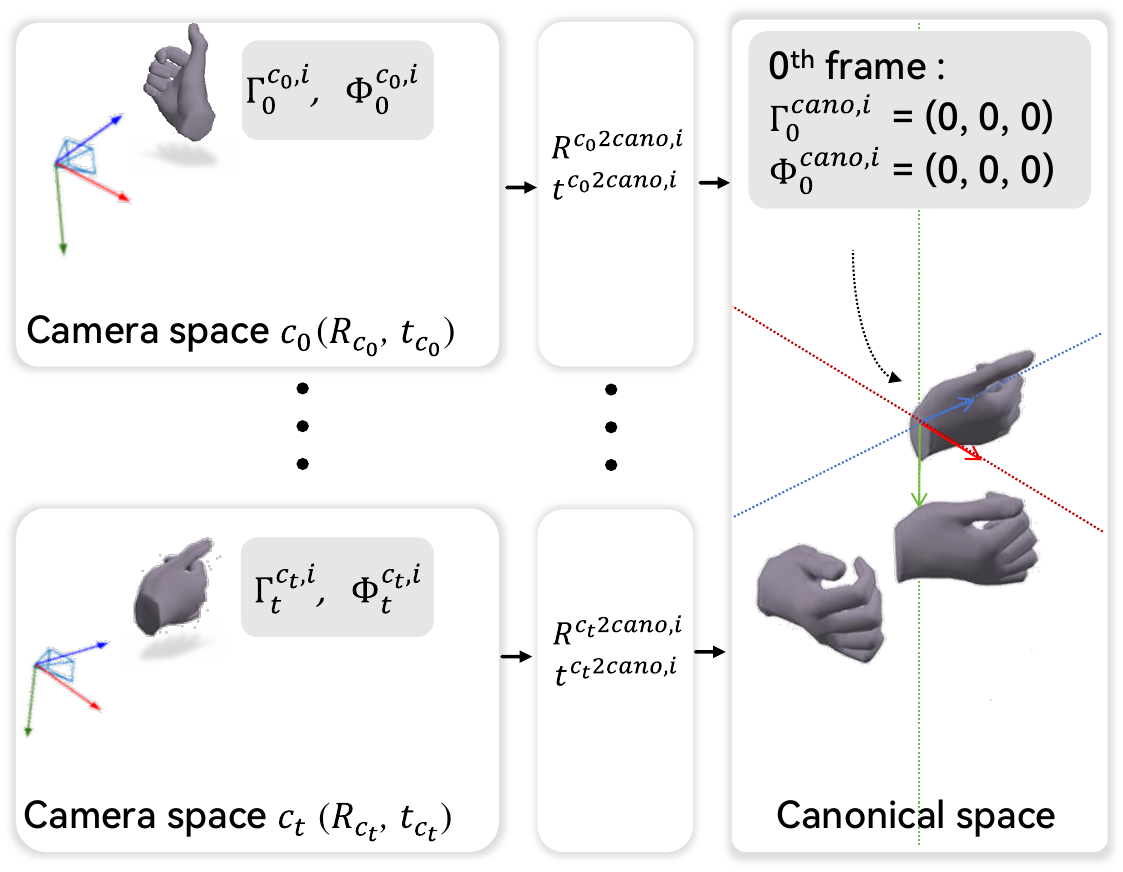}
  \caption{\textbf{Illustration of camera to canonical space transform}. We transform the sequence into canonical space that decouples the hand motion from the dynamic camera and aligns the sequence start state to zero translation and zero rotation.}
  \label{fig:camera2cano}
\end{figure}

We evaluate the reconstructed world-space camera trajectories and hand motions using HOT3D dataset~\cite{banerjee2024introducing}, that contains egocentric videos from Aria glasses accompanied with moving camera trajectories and hand MANO annotations in the world-coordinates. HOT3D is also used to evaluate the infiller network since it provides accurate hand annotations, even when hands are out of the egocentric camera frustum. We use 110 videos as the training set and 27 videos as the validation set.

To evaluate HaWoR we use the following metrics:
\begin{itemize}
    \item \textbf{PA-MPJPE} and \textbf{AUC}. To evaluate 3D hand pose in the camera-frame, we compute the Procrustes-Aligned Mean Per Joint Position Error (PA-MPJPE) measured in millimeters ($mm$) and Area Under the Curve (AUC) to assess the 3D joint accuracy.
    \item \textbf{W-MPJPE} and \textbf{WA-MPJPE} that measure the MPJPE in $mm$, for a sliced sequence of 100-frame segments, after aligning the first frames and aligning the entire trajectories, respectively.
    \item \textbf{RTE}. We evaluate the Root Translation Error (RTE in \%) normalized by the displacement of the hand trajectories after rigid alignment.
    \item \textbf{Accel}. We compute Acceleration error (Accel, in $m/s^2$) that measures the inter-frame smoothness of the reconstructed motion.
    \item \textbf{FID} is the Frechet Inception Distance that calculates the distribution distance between MANO space of the estimated and GT sequence. 
    \item \textbf{ATE} and \textbf{ATE-S}. We compute the Average Trajectory Error (ATE), which uses Procrustes analysis to align the scale to GT. ATS-S is adopted to report the Average Trajectory Error with the estimated scale.
\end{itemize}

\begin{figure}[t]
  \centering
  \includegraphics[width=1\columnwidth]{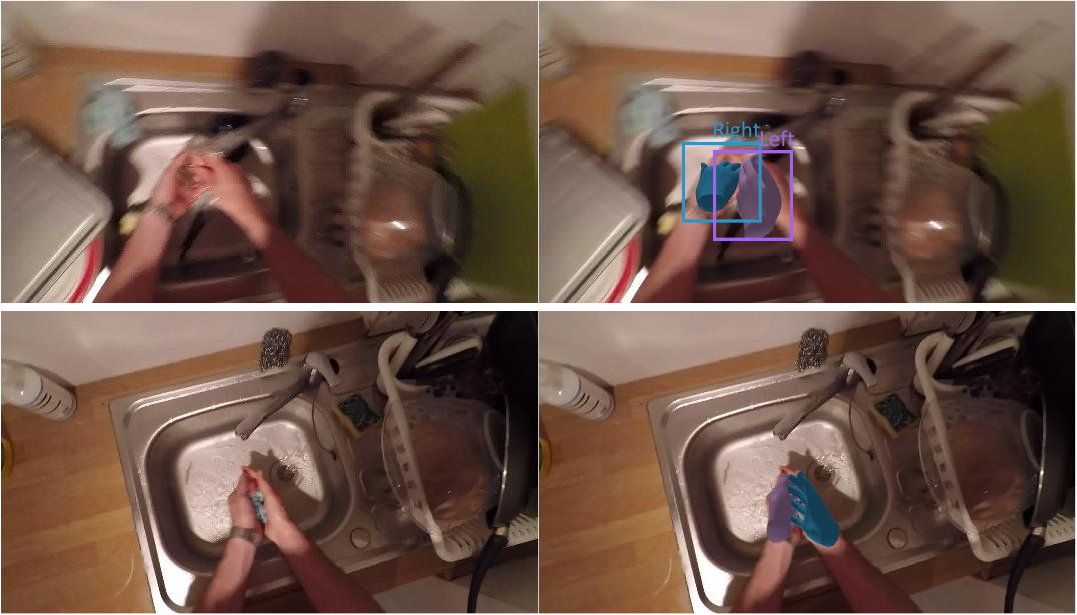}
  \caption{Failure cases of HaWoR in hand motion reconstruction.}
  \label{fig:failcase}
\end{figure}

\section{Limitations}
One limitation of our approach is its reliance on hand-tracking outputs from an off-the-shelf method~\cite{potamias2024wilor}, which can propagate erroneous detections to HaWoR, particularly in cases of tracking identity failures. As illustrated in \cref{fig:failcase}, such issues can lead to reconstruction errors, for example, when left/right hand tracking is incorrect.

It is also important to note that HaWoR models each hand independently, without any inter-penetration constrains. This can cause self-penetrations when the two hands interac, as shown in \cref{fig:failcase}. 
In the future we plan to explore both hand interactions and further constrain the penetration between the two hands. 

\end{document}